\documentclass[10pt,twocolumn,letterpaper]{article}

\usepackage{iccv}
\usepackage{times}
\usepackage{epsfig}
\usepackage{graphicx}
\usepackage{amsmath}
\usepackage{amssymb}

\usepackage[inline]{enumitem}
\usepackage[dvipsnames]{xcolor}
\usepackage{booktabs}
\usepackage{tabularx}
\usepackage{multirow}
\usepackage{colortbl}
\usepackage{subcaption}
\usepackage{float}

\usepackage{algpseudocode,algorithm,algorithmicx}
\usepackage{xspace}
\usepackage{interval}
\usepackage{bm,upgreek}
\usepackage[pagebackref=true,breaklinks=true,letterpaper=true,colorlinks,citecolor=ForestGreen, bookmarks=false]{hyperref}

\usepackage[breaklinks=true,bookmarks=false]{hyperref}

\iccvfinalcopy 

\def\iccvPaperID{****} 

\begin{document}
\newcommand{\ImageNetCls}{{\scshape ImageNet-Cls}\xspace}
\newcommand{\ImageNetLoc}{{\scshape ImageNet-Loc}\xspace}
\newcommand{\COCO}{{\scshape Coco}\xspace}
\newcommand{\OpenImages}{{\scshape OpenImages}\xspace}
\newcommand{\SUN}{{\scshape Sun-397}\xspace}
\newcommand{\Cal}{{\scshape Caltech-256}\xspace}
\newcommand{\PASCALVOC}{{\scshape Pascal-Voc}\xspace}
\newcommand{\Flower}{{\scshape Oxford-102 Flowers}\xspace}
\newcommand{\cmt}[1]{{\footnotesize\textcolor{red}{#1}}}
\newcommand{\todo}[1]{\cmt{(TODO: #1)}}

\title{An Analysis of Pre-Training on Object Detection}

\author{Hengduo Li \quad Bharat Singh \quad Mahyar Najibi \quad Zuxuan Wu \quad Larry S. Davis\\
\\
University of Maryland, College Park\\
{\tt\small \{hdli,bharat,najibi,zxwu,lsd\}@cs.umd.edu}
}

\maketitle

\begin{abstract}
We provide a detailed analysis of convolutional neural networks which are pre-trained on the task of object-detection. To this end, we train detectors on large datasets like OpenImagesV4, ImageNet Localization and COCO. We analyze how well their features generalize to tasks like image classification, semantic segmentation and object detection on small datasets like PASCAL-VOC, Caltech-256, SUN-397, Flowers-102 etc. Some important conclusions from our analysis are---1) Pre-training on large detection datasets is crucial for fine-tuning on small detection datasets, especially when precise localization is needed. For example, we obtain 81.1\% mAP on the PASCAL-VOC dataset at 0.7 IoU after pre-training on OpenImagesV4, which is 7.6\% better than the recently proposed DeformableConvNetsV2 which uses ImageNet pre-training. 2) Detection pre-training also benefits other localization tasks like semantic segmentation but adversely affects image classification. 3) Features for images (like avg. pooled Conv5) which are similar in the object detection feature space are likely to be similar in the image classification feature space but the converse is not true. 4) Visualization of features reveals that detection neurons have activations over an entire object, while activations for classification networks typically focus on parts. Therefore, detection networks are poor at classification when multiple instances are present in an image or when an instance only covers a small fraction of an image.
\end{abstract}

\section{Introduction}

For several computer vision problems like object detection, image segmentation and image classification, pre-training on large scale datasets is common~\cite{fcn, rcnn, decaf}. This is because it leads to better results and faster convergence~\cite{places, tl_quoc, decaf, instagram, rethinking}. However, the effect of pre-training in computer vision is often evaluated by training networks for the task of {\em image classification}, on datasets like ImageNet \cite{imgnet}, Places \cite{places}, JFT \cite{jft}, Instagram \cite{instagram} \etc, but rarely for object detection. It can be argued that the task of object detection subsumes image classification, so a network good at object detection should learn richer features than one trained for classification. After all, this network has access to an orthogonal semantic information, like the spatial extent of an object. However, it can also be argued that forcing a network to learn position sensitive information may affect its spatial invariance properties which help in recognition. To this end, we provide a comprehensive analysis which compares pre-training CNNs on object detection and image classification. 

We pre-train a network on the OpenImagesV4 \cite{openimages} (hereafter referred to as \OpenImages) dataset on the object detection task and fine-tune it on tasks like semantic segmentation, object detection and classification on datasets like \PASCALVOC~\cite{pascal}, \COCO~\cite{coco}, \Cal~\cite{cal256}, \SUN~\cite{sun397}, and \Flower~\cite{flowers}. For a stronger evaluation, we also pre-train on the ImageNet classification dataset~\cite{imgnet} with bounding-box annotations on 3,130 classes~\cite{rfcn3k} (hereafter referred to as \ImageNetLoc as opposed to \ImageNetCls for ImageNet Classification dataset without bounding boxes) and the \COCO dataset~\cite{coco} which helps us in evaluating the importance of the number of training samples. We then design careful experiments to understand the differences in properties of features which emerge by pre-training on detection \vs classification. 

Our experimental analysis reveals that pre-training on object detection can improve performance by more than 5\% on \PASCALVOC for object detection (especially at high IoUs) and 3\% for semantic segmentation. However, detection features are significantly worse at performing classification compared to features from \ImageNetCls pre-trained networks ($\sim$ 8\% on \Cal). We also find that if features (like average pooled \texttt{Conv5}) are similar in the object detection feature space, they are likely to be similar in the image classification feature space, but the converse is not true. Visualization of activations for object detection shows that they often cover the entire extent of an object, so are poor at recognition when an object is present in a small part of an image or when multiple instances are present. 

\section{Related Work}

\textbf{Large Scale Pre-training} 
The initial success of deep learning in computer vision can be largely attributed to transfer learning. ImageNet pre-training was crucial to obtain improvements over state-of-the-art results on a wide variety of recognition tasks such as object detection~\cite{fpn, ssd, yolo, snip, rfcn, fpn}, semantic segmentation~\cite{deeplab, maskrcnn, deeplabv3, fcn, pspnet}, scene classification~\cite{places, scene_1}, action/event recognition~\cite{action_0, action_1, action_2, action_3} \etc. Due to the importance of pre-training, the trend continued towards collecting progressively larger classification datasets such as JFT~\cite{jft}, Places~\cite{places} and Instagram~\cite{instagram} to obtain better performance. While the effect of large-scale classification is extensively studied \cite{sharif2014cnn,decaf}, there is little work on understanding the effect of pre-training on object detection.

\textbf{Transfer Learning} The transferability of pre-trained features has been well studied~\cite{tl_factor, trans_01, trans_02, tl_quoc, tl_cui, jft}. For example, ~\cite{tl_factor} measured the similarity between a collection of tasks with ImageNet classification; ~\cite{tl_cui} studied how to transfer knowledge learned on large classification datasets to small fine-grained datasets; ~\cite{tl_quoc} addressed relationship among ImageNet pre-training accuracy, transfer accuracy and network architecture; ~\cite{taskonomy} proposed a computational approach to model relationships among visual tasks of various abstract levels and produced a computational taxonomic map. However, the visual tasks in ~\cite{taskonomy} did not involve object detection although object detection is one of the few tasks other than image-classification for which large-scale pre-training can be performed. We study the transferability, generalizability, and internal properties of networks pre-trained for object detection.

\textbf{Understanding CNNs} Towards understanding the superior performance of CNNs on complex perceptual tasks, various qualitative~~\cite{vis_simonyan, vis_cam, vis_gradcam, vis_syn_3, vis_deconv, vis_syn_nips, vis_guidedbp, vis_syn_4, vis_syn_2} and quantitative~~\cite{shape_1, shape_2, texture_1, texture_2} approaches have been proposed. A number of previous works explain the internal structure of CNNs by highlighting pixels which contribute more to the prediction using gradients~\cite{vis_simonyan}, Guided BackPropogation~\cite{vis_guidedbp, vis_gradcam}, deconvolution~\cite{vis_deconv}, \etc. Other methods adopt an activation maximization based approach and synthesize the preferred input for a network neuron~\cite{vis_syn_nips, vis_syn_2, vis_syn_3, vis_syn_4}. Attempts have also been made to interpret the properties of CNNs empirically by investigating what it learns and is biased towards. While~\cite{shape_1, shape_2} suggest that deep neural networks implicitly learn representations of shape, recent work~\cite{texture_1, texture_2} indicates that CNNs trained for image classification task are biased towards texture. 
~\cite{texture_1} further indicates the advantage of a shape-based representation by training CNNs on a stylized version of ImageNet.

\textbf{Training From Scratch} While most modern detectors are pre-trained on the ImageNet classification dataset~\cite{fpn, snip, rfcn, fpn, ssd, yolo}, effort has also been made to deviate from the conventional paradigm and train detectors from scratch~\cite{dsod, cornernet}. ~\cite{dsod} proposed a set of design principles to train detector from scratch. ~\cite{rethinking} demonstrated that with a longer training schedule, detectors trained from scratch can be as good as ImageNet pre-trained models on large datasets (like \COCO). However, pre-training is still crucial when the training dataset is small (like \PASCALVOC).

\section{Discussion}
A detailed analysis of detection pre-training is lacking in the existing literature. This is primarily because \COCO \cite{coco} is still small compared to \ImageNetCls (by 10$\times$ images, 10$\times$ categories), so there is an unknown variable about the scale of the dataset. While \ImageNetLoc also contains bounding-boxes for objects, detection in it is not challenging as images typically only contain a single object (which are often large, making localization trivial in many cases), so in this case, it is unclear if the network is learning instance level features. Recently, due to a massive data collection effort, a new dataset called \OpenImages was released which contains bounding-box annotations for 15 million instances and close to 2 million images. This is the first dataset which provides an orthogonal semantic information at the scale of ImageNet. Therefore, it allows us to fairly compare networks pre-trained on large scale object detection with large scale image classification when fine-tuning on standard computer vision datasets in which the number of annotations is lower by one or two orders of magnitude.

\section{Analysis}
We perform pre-training on multiple detection datasets and compare it with \ImageNetCls pre-training for different computer vision tasks like object detection, image classification and semantic segmentation. For detection pre-training, our experimental setup is as follows. All our detection networks are pre-trained first on \ImageNetCls. They are then trained on detection datasets like \OpenImages~\cite{openimages}, \ImageNetLoc~\cite{imgnet, rfcn3k} and \COCO~\cite{coco}. 
The SNIPER~\cite{sniper} detector is trained on all the datasets. We use multiple pre-training datasets for two reasons - 1) To thoroughly evaluate our claims about pre-training for the detection task 2) Since the datasets contain a different number of classes and training examples, it also provides an indication of the magnitude of improvement one can expect by pre-training on detection datasets of different sizes. 

\begin{table}[h]
    \centering
    \resizebox{\linewidth}{!}{
    \begin{tabular}{@{}*{4}c@{}}
    \toprule
     \textbf{Dataset} & \textbf{\#Class} & \textbf{\#Images} & \textbf{\#Objects} \\
     \midrule
      \ImageNetCls~\cite{imgnet} & 1000 & 1.28M & -\\
     \midrule
      \OpenImages~\cite{openimages} & 500 & 1.74M & 14.6M \\
      \ImageNetLoc~\cite{imgnet, rfcn3k} & 3,130 & 0.92M & 1.06M \\
      \COCO~\cite{coco} & 80 & 0.14M  & 0.89M \\ 
     \midrule
     \Cal~\cite{cal256} & 257 & 15.4K/15.2K & -\\
     \SUN~\cite{sun397} & 397 & 19.9K/19.9K & -\\
     \Flower~\cite{flowers} & 102 & 2.0K/6.1K & -\\
     \midrule
     \PASCALVOC Det~\cite{pascal} & 20 & 16.6K/5.0K & 40.1K/12.0K \\
     \PASCALVOC Seg~\cite{pascal} & 21 & 10.6K/1.4K & -/-\\
     \bottomrule
     \end{tabular}}
 \caption{Source and target datasets examined. $x/y$ denotes $x$ for training set and $y$ for evaluation set.}
 \label{table:dataset_stat}
\end{table}
\textbf{Datasets} Here we briefly introduce the target datasets used in our fine-tuning experiments. For the \textit{object detection} task, we fine-tune on the \PASCALVOC dataset~\cite{pascal}. We use the VOC 07+12 trainval set for training and the VOC 07 test set for evaluation. For the \textit{semantic segmentation} task,  we follow ~\cite{dcnv1, vocseg1, vocseg2, vocseg3} and use VOC 2012 plus additional annotations provided in~\cite{vocseg_data}. For \textit{image classification}, we fine-tune on \Cal~~\cite{cal256}, \SUN~~\cite{sun397} and \Flower~~\cite{flowers}. We use the trainval and test sets directly in \Cal and \Flower; for \SUN we follow~\cite{tl_quoc} and use the first split for training and evaluation. The number of classes, images and instances in each of these datasets are mentioned in Table \ref{table:dataset_stat}.

\textbf{Architecture} We briefly describe the architecture of the two detection heads (Faster-RCNN and R-FCN) which are used for training. On \OpenImages detector after \texttt{Conv5} (2048,14,14) we have the following layers: \texttt{ConvProj} (256,14,14), \texttt{FC1} (1024), \texttt{FC2} (1024), \texttt{Output} (501), \texttt{Regression} (4). A fully connected layer projects the (256,14,14) blob to a 1024 dimensional vector, thus spatial information is preserved for the blob. The \texttt{Output} (501) and \texttt{Regression} (4) layers are connected to \texttt{FC2}. The same architecture is used for the \COCO detector, except that the \texttt{Output} layer has 81 dimensions. For the \ImageNetLoc detector, the architecture is as described in ~\cite{rfcn3k}. In this architecture, classification and detection are decoupled and performed independently. For classification, \texttt{Conv5} features are average pooled and a fully connected layer projects these 2048 dimensional features to a 1024 vector, on which a 3130 dimensional classifier is applied. Detection is performed using a R-FCN head on the \texttt{Conv5} features which are first projected to 1024 dimensional features.  

\subsection{Object Detection}
\textbf{Baseline Configuration and Results} For our object detection experiments, we train our detectors (SNIPER with ResNet-101) on 3 datasets:  \OpenImages, \COCO and \ImageNetLoc. Our \OpenImages model obtains 45\% mAP (at 0.5 overlap) on the validation set. It is trained at 2 scales, (480,512) and (768, 1024) without negative chip mining. Inference is also performed at these two scales only. For the \COCO model, training and inference is performed at 3 scales (480,512), (800, 1280) and (1400,2000) and the detector obtains an mAP of 46.1\% (\COCO metric) on the test-dev set. The \ImageNetLoc model obtains 37.4\% mAP (at 0.5 overlap) on the ImageNet \emph{Detection} dataset (not \ImageNetLoc). This detector was only trained at a single scale of (512,512) on \ImageNetLoc without any negative chip mining. Inference is also performed only at a scale of (512,512) as compared to others, this dataset contains relatively bigger objects.

\begin{table}[!b]
\centering
\resizebox{\linewidth}{!}{
\begin{tabular}{@{}*{3}c@{}}
  \toprule
    \textbf{Method / Pre-trained Dataset} &  mAP@\textbf{0.5} & mAP@\textbf{0.7} \\
 \midrule
    DCNv1~\cite{dcnv1} & 81.9 & 68.2 \\
    DCNv2~\cite{dcnv2} & 84.9 & 73.5 \\
 \midrule
    \ImageNetCls~\cite{imgnet} & 84.6 & 76.3 \\
    \ImageNetLoc~\cite{imgnet, rfcn3k} & 86.5 & 80.0 \\
    \COCO~\cite{coco} & 86.8 & 80.7 \\
    \OpenImages~\cite{openimages} & \textbf{86.8} & \textbf{81.1} \\
 \bottomrule
 \end{tabular}}
 \caption{Baseline and our results on \PASCALVOC 2007~\cite{pascal} object detection dataset.}
 \label{table:det_result}
\end{table}

\begin{figure}[t!]
  \includegraphics[width=\linewidth]{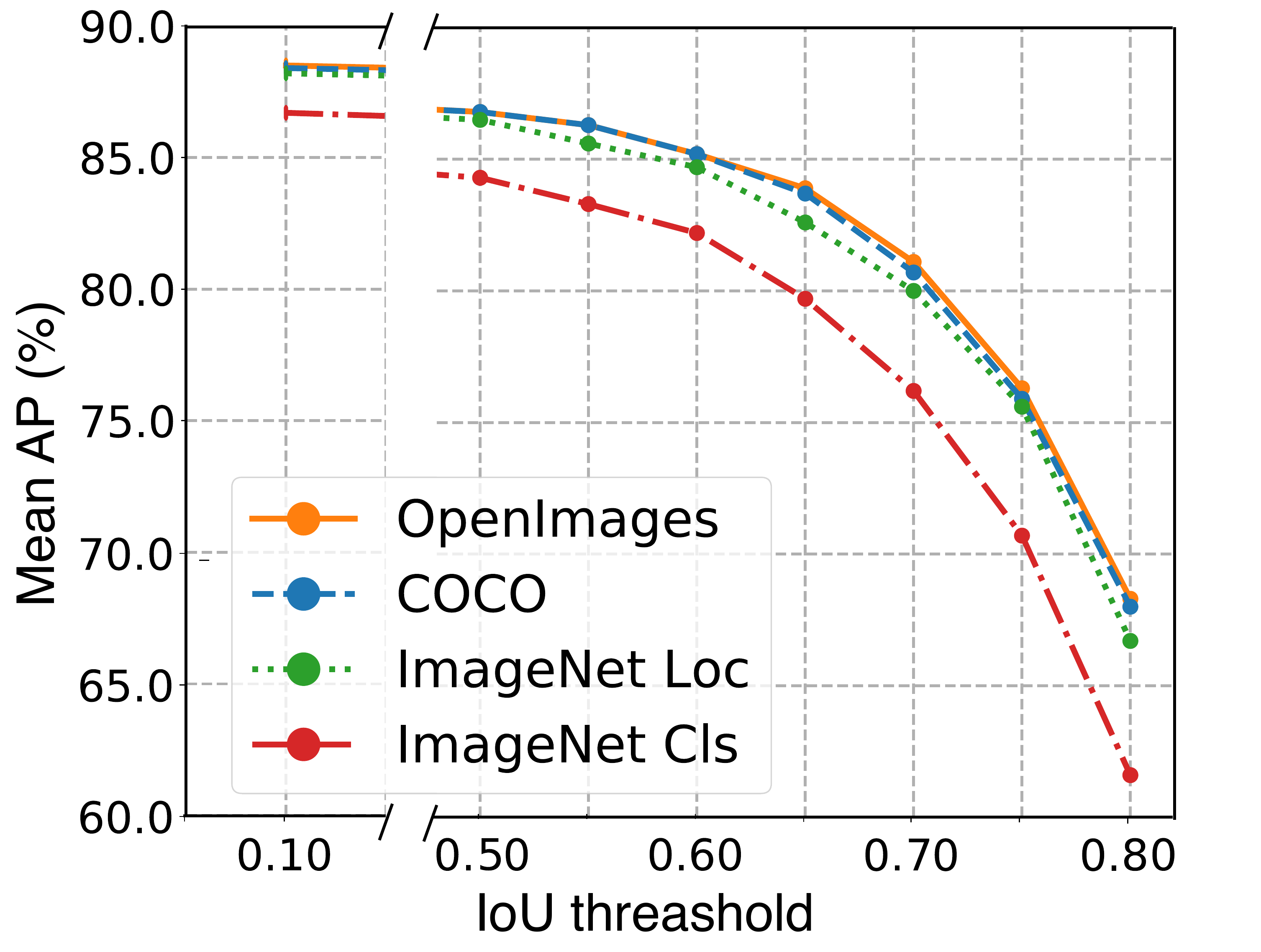}
  \caption{Detection performance (mAP \%) at different IoUs on \PASCALVOC 2007~\cite{pascal} test set of detectors pre-trained on different datasets. Typically, localization errors are ignored at $0.1$ IoU threshold.}
  \label{fig:det_plot}
\end{figure}

\begin{figure*}[t!]
  \includegraphics[width=\linewidth]{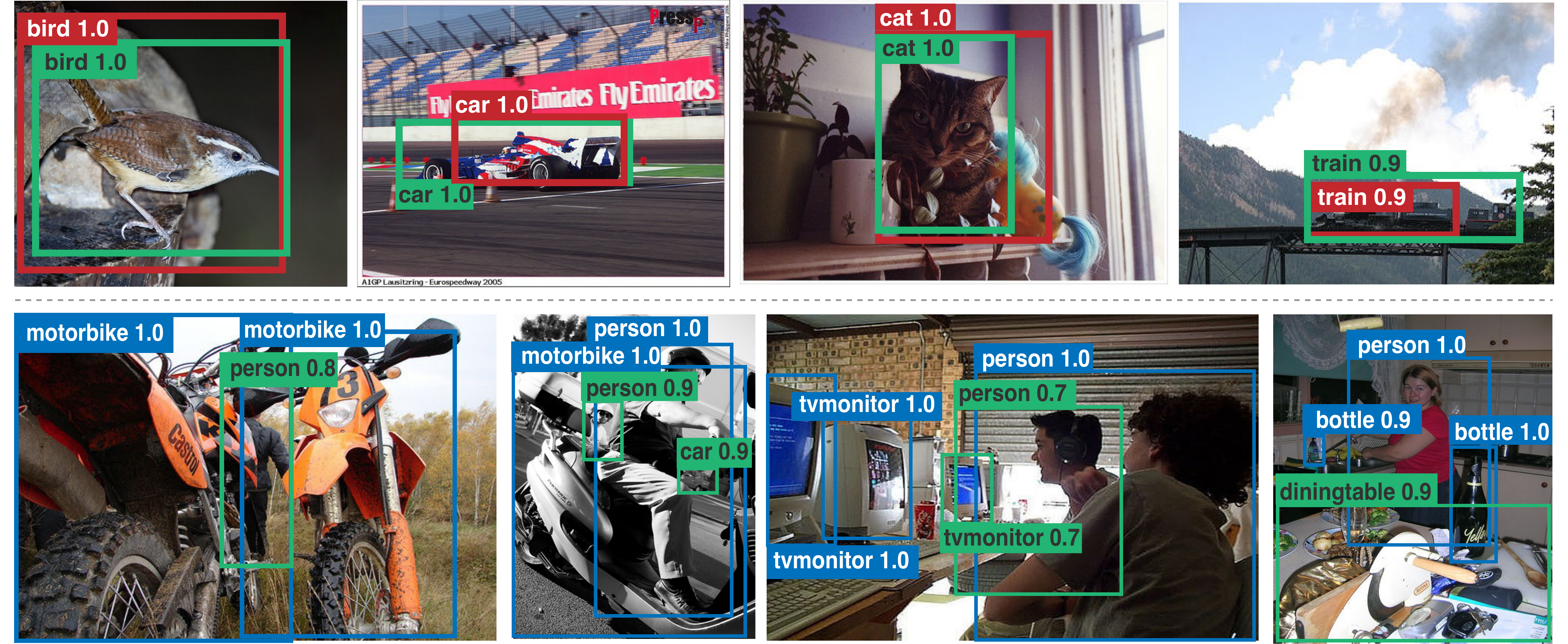}
  \caption{Qualitative results on \PASCALVOC 2007 from detectors pre-trained on \ImageNetCls~\cite{imgnet} and \OpenImages~\cite{openimages} \textbf{Above}: \OpenImages pre-trained detector shows better localization ability. Green and red boxes are from \OpenImages pre-trained and \ImageNetCls pre-trained detectors respectively. \textbf{Below}: \OpenImages pre-trained detector handles occlusion cases better. Blue boxes are correct predictions from both detectors while green boxes are occluded objects successfully detected only by the \OpenImages pre-trained detector.}
  \label{fig:det_quals}
\end{figure*}
\textbf{Fine-tuning on \PASCALVOC} We fine-tune these pre-trained models on \PASCALVOC~\cite{pascal} using the same set of scales as \COCO for both training and inference. Detection heads of the models pre-trained on detection datasets are re-initialized before fine-tuning.  Following \cite{sniper}, we train the RPN for 2 epochs first for negative chip mining. Then training is performed for 7 epochs with learning rate step-down at the end of epoch 5. Horizontal flipping is used as data augmentation.  The results are shown in Table \ref{table:det_result}. 
As a reference, the recently proposed Deformable ConvNet-V2~\cite{dcnv2} obtains 73.5\% at an overlap of 0.7 while the \OpenImages/\COCO/ImageNet-3k models obtain 81.1\%, 80.7\% and 80\% mAP at 0.7 overlap. Our baseline network which is only trained on \ImageNetCls obtains an mAP of 76.3\%. Thus, pre-training on larger detection datasets can improve performance on PASCAL by as much as 4.8\% at an overlap of 0.7. However, such large improvements do not translate to lower overlap thresholds. For example, the difference in mAP between \ImageNetCls and the \OpenImages model at an overlap of 0.5 is only 2.2\%. We plot the mAP for all the detection models at different overlap thresholds in Fig \ref{fig:det_plot}.
This clearly shows that pre-training for detection helps to a large extent in improving localization performance on PASCAL. We also observe this phenomenon when we use the \OpenImages model to fine-tune on the \COCO dataset. For example, the performance at an overlap of 0.5 on \COCO with \ImageNetCls pre-training is 67.5 and at 0.75 it is 52.2. When \OpenImages pre-training is used, the performance at 0.5 improves by 0.7\%, but results at 0.75 improve by 1.4\%.

\textbf{Pre-training helps at Higher IoU} While the \COCO result that mAP at 0.75 improves more than mAP at 0.5 after fine-tuning from an \OpenImages model was presented in SNIPER \cite{sniper}, here we show that this is indeed a systematic pattern which is observed when pre-training is performed on large scale detection datasets. When the size of the detection dataset is small (like \PASCALVOC), localization at higher overlap thresholds can significantly benefit from pre-training on large detection datasets. Another pattern we observe here is that the number of samples in the pre-trained dataset did not affect the fine-tuning performance to a large extent (differences are within 1\%). The important factor was whether the network was pre-trained on a reasonably large detection dataset ($>$ 1M training instances) or not.

\begin{figure*}[t!]
  \includegraphics[width=\linewidth]{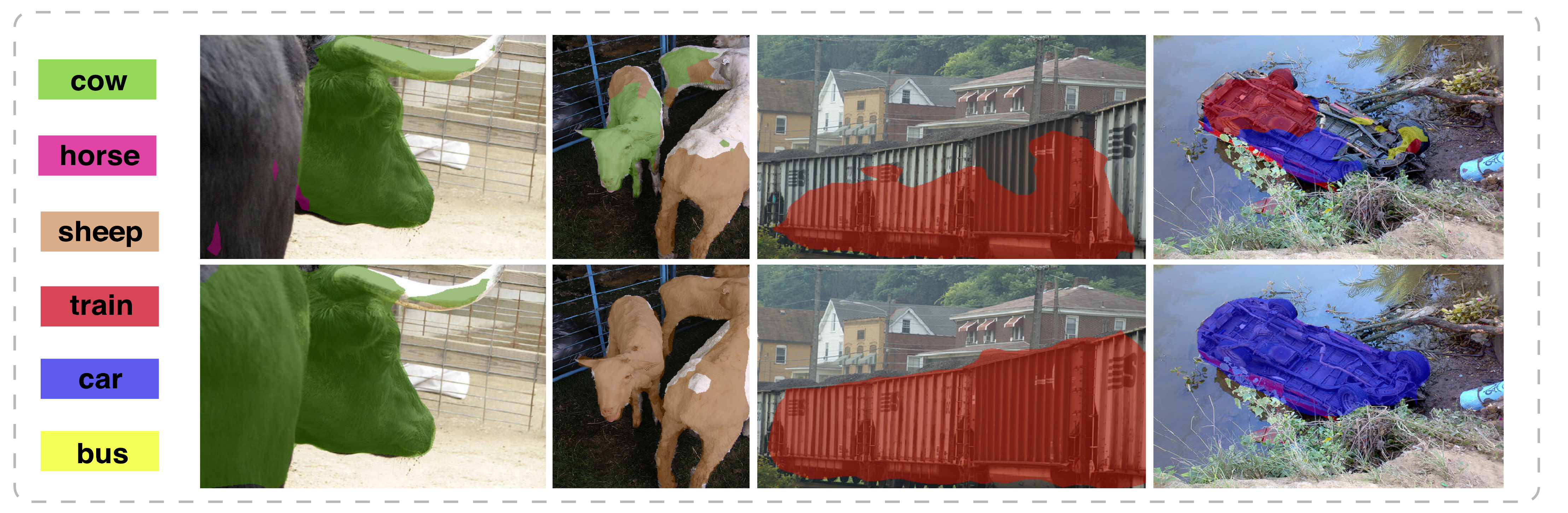}
  \caption{Qualitative results of semantic segmentation from networks pre-trained on \ImageNetCls~\cite{imgnet} (\textbf{Above}) and \ImageNetLoc~\cite{imgnet, rfcn3k} (\textbf{Below}). The \ImageNetLoc model is better at covering entire objects while the classification pre-trained model is more likely to mis-classify pixels on some parts of an object.} \label{fig:seg_quals}
\end{figure*}

\textbf{Qualitative Results and Error Analysis} 
We show qualitative results on the \PASCALVOC dataset for \OpenImages and \ImageNetCls pre-training. Fig \ref{fig:det_quals} shows that localization for the \OpenImages model is better. Following \cite{coco}, we evaluate detectors pre-trained on different aforementioned datasets at different IoU thresholds including $0.1$, where localization errors are typically ignored. The small gap between mAP$@$0.1 and higher IoUs like 0.5 indicates that large localization errors are rare. We also observe in Fig \ref{fig:det_quals} that the \OpenImages model handles occlusion cases better. To further verify the observation on performance improvement under occlusion, we also analyze the errors using the object detection analysis tools in~\cite{derek, coco}. Quantitative results are mentioned in Table~\ref{table:occ_stat} which demonstrate that the \OpenImages network is indeed better under occlusion.

\begin{table}[!h]
\centering
\resizebox{\linewidth}{!}{
 \begin{tabular}{@{}*{3}c@{}}
  \toprule
    \% \textbf{missed object} &  \textbf{occluded Low} &  \textbf{occluded Medium} \\
 \midrule
    \ImageNetCls~\cite{imgnet} & 14.7\% & 15.7\%  \\ 
    \OpenImages~\cite{openimages} & \textbf{10.1\%} & \textbf{10.8\%} \\
 \bottomrule
 \end{tabular}}
 \caption{Percentage of missed objects under low and medium occlusion levels in the \PASCALVOC 2007~\cite{pascal} test set. Results are obtained using the object detection analysis tool in~\cite{derek}}
 \label{table:occ_stat}
\end{table}

\begin{figure}[b!]
  \includegraphics[width=\linewidth]{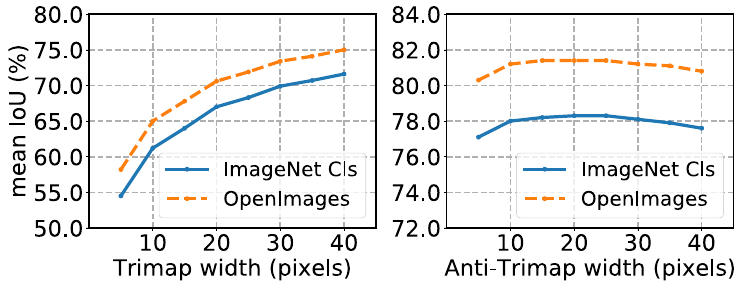}
  \caption{Results of Trimap (\textbf{left}) and Anti-Trimap(\textbf{right}) experiments. Segmentation performance on pixels inside an $x$-pixel-wide trimap band (near object boundary) and outside the trimap band (away from object boundary) are evaluated respectively.}
  \label{fig:trimap}
\end{figure}

\subsection{Semantic Segmentation}
\textbf{Baseline Configuration and Results} We fine-tune detection networks for the semantic-segmentation task on \PASCALVOC 2012.
Following ~\cite{dcnv1}, we use Deformable ConvNets~\cite{dcnv1} as our backbone in DeepLab ~\cite{deeplab} throughout our experiments. In training and inference, images are resized to have a shorter side of 360 pixels while keeping the larger side less than 600 pixels. Baseline results are shown in Table \ref{table:seg_result}.

\textbf{Detection Pre-Training Helps Segmentation} The results after fine-tuning are shown in Table \ref{table:seg_result}. These results show that networks which are trained for object detection obtain a 3\% improvement in performance compared to image classification. We evaluate this for the OpenImages dataset and also for \ImageNetLoc dataset.

\begin{figure*}[t!]
  \includegraphics[width=\linewidth]{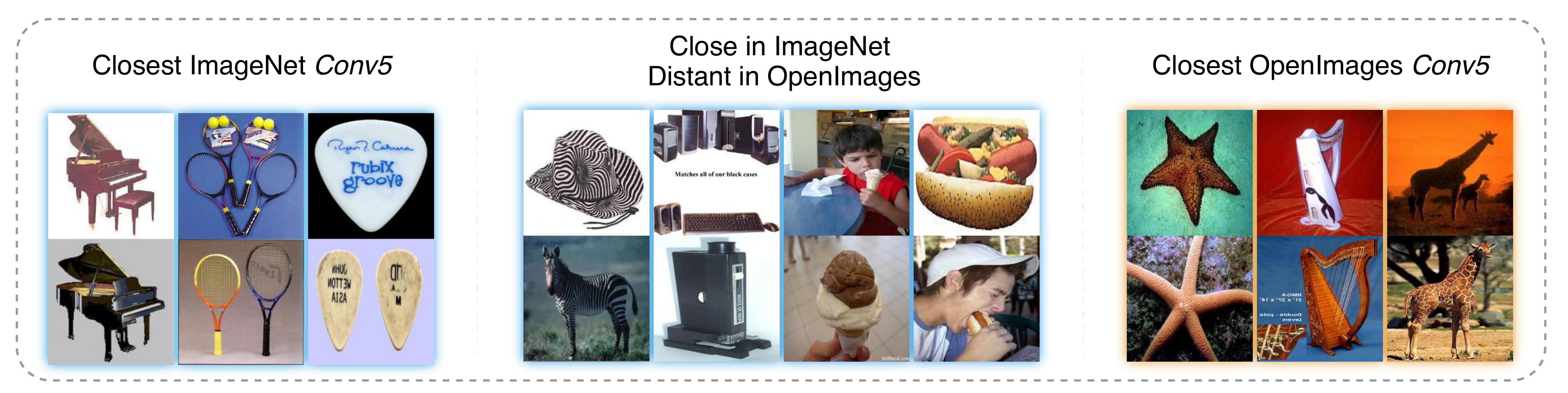}
  \caption{Qualitative results of our feature space analysis. We extract \texttt{Conv5} features of \Cal~\cite{cal256} dataset with networks pre-trained on \ImageNetCls~\cite{imgnet} and \OpenImages~\cite{openimages} without fine-tuning, then qualitatively analyze the two networks in feature space based on $\ell_2$ distance. \textbf{Left/Right}: Image pairs that are closest in feature space of \ImageNetCls/\OpenImages pre-trained network (using inner-product). \textbf{Middle}: Image pairs that are close in \ImageNetCls pre-trained network but distant in \OpenImages pre-trained network.}
  \label{fig:cls_quals}
\end{figure*}

\textbf{Error Analysis} We also perform experiments to understand where these improvements occur. Specifically, we study if the improvements from detection pre-trained networks are due to better segmentation at boundary pixels or not. For this we evaluate the accuracy at boundary pixels with the ``trimap experiment''~\cite{trimap_1, trimap_2, deeplab, deeplabv3} and non-boundary pixels called the ``anti-trimap experiment''. The boundary pixels are obtained by applying morphological dilation on the ``void'' labeled pixels which often occurs at object boundaries.

We perform two types of evaluations. 1) Accuracy at pixels which are within a distance $x$ from an object boundary 2) Accuracy at pixels of an object or background, not in (1). The first evaluation compares the accuracy at boundary pixels while the second one compares the accuracy for pixels which are not at the boundary. The results for these experiments are shown in Fig. \ref{fig:trimap} (using \OpenImages). These results show that the gap in performance remains the same as the size of boundary pixels is increased (instead of reducing, if one model was more accurate at boundaries). The same is true for pixels which are far away from the boundary. There is still a significant gap in performance for pixels which are not at the boundary. Thus, from these experiments, it is clear that improvement in performance is not due to better classification at boundary pixels.

\begin{table}[!t]
\centering
\resizebox{0.73\linewidth}{!}{
\footnotesize
 \begin{tabular}{@{}*{2}c@{}}
  \toprule
    \textbf{Method / Pre-trained Dataset} &  \textbf{mIoU} \\
 \midrule
    DCNv1~\cite{dcnv1} & 75.2  \\
 \midrule
    \ImageNetCls~\cite{imgnet} & 75.7  \\
    \ImageNetLoc~\cite{imgnet, rfcn3k} & 78.3 \\
    \OpenImages~\cite{openimages} & \textbf{78.6} \\
 \bottomrule
 \end{tabular}}
 \caption{Baseline and our fine-tuning results on \PASCALVOC 2012~\cite{pascal} semantic segmentation dataset.}
 \label{table:seg_result}
\end{table}

\textbf{Qualitative and Semantic Analysis} We provide some qualitative examples for segmentation predictions in Fig \ref{fig:seg_quals} (using \ImageNetLoc and \ImageNetCls). From these examples, we find that the network pre-trained on classification is unable to cover entire objects as it is weak at understanding instance boundaries - like in the case of the cow in Fig \ref{fig:seg_quals}. Detection pre-training provides a better prior about the spatial extent of an instance which helps in recognizing parts of an object. It also helps more for object classes like sheep (+7.5\%), cow (+6.5\%), dining-table (+5.6\%). These classes typically have a multi-modal distribution in appearance (like color and shape distribution). On the other hand, classes like Potted Plant which have a consistent shape and appearance, obtain no improvement in performance when detection pre-training is used.

\subsection{Image Classification} \label{Image Classification}
We also compare the effect of pre-training for image classification by evaluating multiple pre-trained detection backbones like \ImageNetLoc and \COCO apart from \OpenImages. Diverse classification datasets like \Cal, \SUN and \Flower are considered. Apart from fine-tuning for image classification, we also evaluate off-the-shelf features from detection and classification backbones.

\textbf{Fine-Tuning on Classification} Results for fine-tuning different pre-trained networks on classification datasets are shown in Table \ref{table:cls_ft_result}. These results show that pre-training on \ImageNetCls outperforms \ImageNetLoc, \OpenImages, and \COCO by a significant margin on all three classification datasets. Therefore, pre-training for object detection hurts performance for image classification. It is a bit counter-intuitive that a network which also learns about the spatial extent of an object is worse at performing image classification. To get a better understanding of the possible reasons, we evaluate features which are extracted from the pre-trained image classification networks without any fine-tuning.

\begin{table}[!b]
\centering
 \resizebox{\linewidth}{!}{
 \begin{tabular}{@{}*{4}c@{}}
  \toprule
    \textbf{Pre-trained Dataset} &  \Cal~\cite{cal256} & \SUN~\cite{sun397} & \Flower~\cite{flowers} \\
 \midrule
    \ImageNetLoc~\cite{imgnet, rfcn3k} & 82.3 & 58.3 & 90.9 \\
    \COCO~\cite{coco} & 79.8 & 57.8 & 91.4 \\
    \OpenImages~\cite{openimages} & 82.2 & 59.5 & 92.6 \\
    \ImageNetCls~\cite{imgnet} & \textbf{86.3} & \textbf{61.5} & \textbf{95.0} \\
 \bottomrule
 \end{tabular}}
 \caption{Results (Top-1 accuracy) for fine-tuning different pre-trained networks on classification datasets..}
 \label{table:cls_ft_result}
\end{table}

\textbf{\texttt{Conv5} features} We average pool the \texttt{Conv5} features extracted from networks pre-trained on \OpenImages and \ImageNetCls. Then we add a linear classifier followed by a softmax function to perform image classification. The results for different datasets are presented in Table \ref{table:cls_conv5_result}. This shows that without fine-tuning, there exists a large performance gap between the features which are good for object detection \vs those which are trained for the task of image classification. The performance of the average pooled \texttt{Conv5} features of \ImageNetLoc and \OpenImages pre-trained networks is the same for classification on \Cal. For \COCO, the performance drops further by 2\%, possibly because of the smaller number of classes in object detection.
\begin{figure}[t]
  \includegraphics[width=\linewidth]{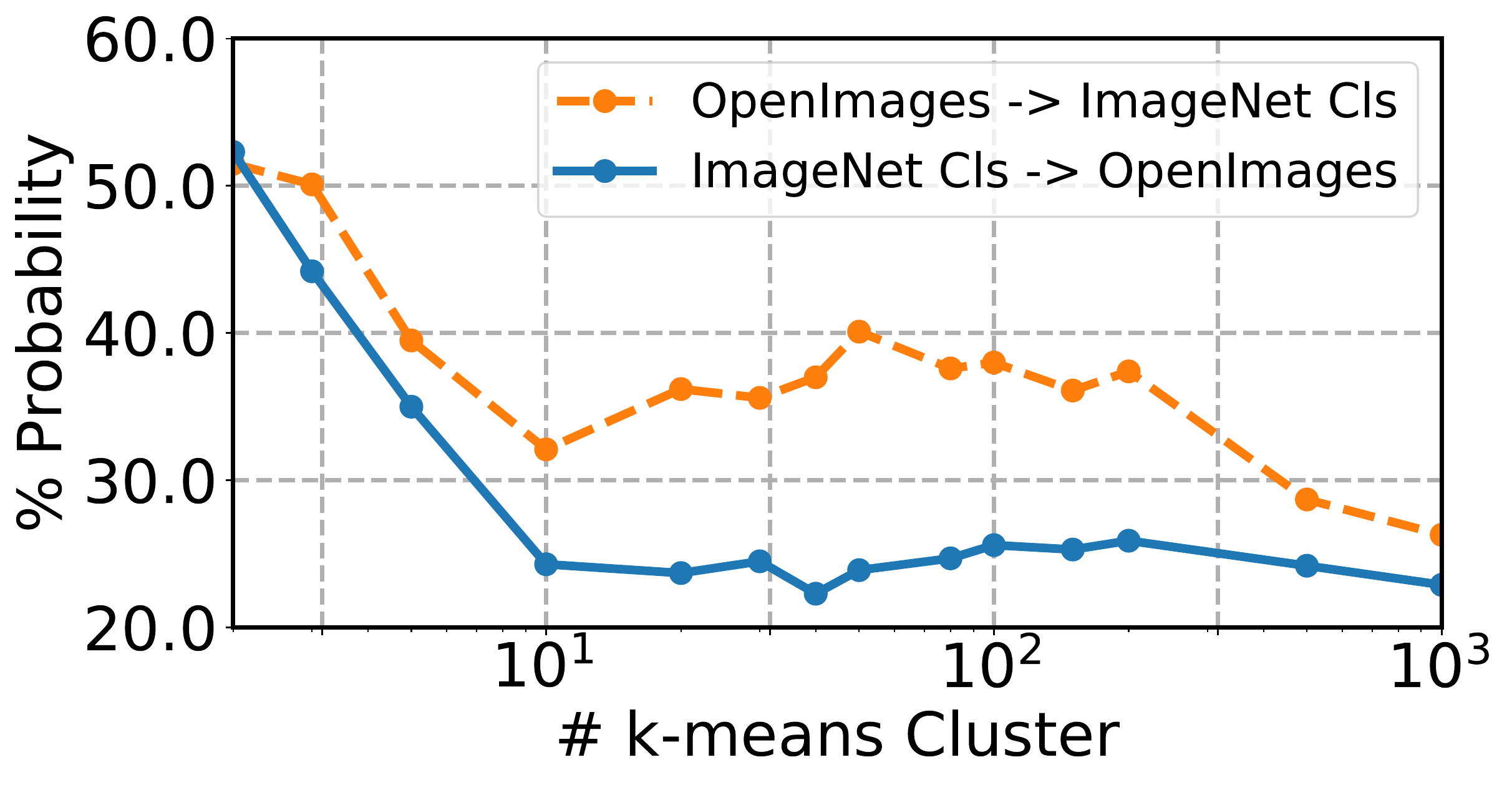}
  \caption{A pair of features within a cluster in one feature space is randomly selected and checked whether it is also assigned to the same cluster in the other feature space. We sample 100,000 pairs and measure the probability of success. The statistics indicate that similar features in \OpenImages space are more likely to be similar in \ImageNetCls space but if features are similar in \ImageNetCls space, it is less likely for them to be close in the \OpenImages space.}
  \label{fig:feat_subsume}
\end{figure}

\begin{table}[!t]
\centering
\resizebox{0.95\linewidth}{!}{
 \begin{tabular}{@{}*{3}c@{}}
 \toprule
    \textbf{Pre-trained Dataset} &  \ImageNetCls~\cite{imgnet, rfcn3k} & \OpenImages~\cite{openimages} \\
 \midrule
    \Cal~\cite{cal256} & \textbf{84.7} & 76.7\\
    \SUN~\cite{sun397} & \textbf{57.3} & 51.1\\
    \Flower~\cite{flowers} & \textbf{87.4} & 83.1\\
 \bottomrule
 \end{tabular}}
 \caption{Linear classification results (Top-1 Accuracy) using \texttt{Conv5} features from \ImageNetCls and \OpenImages pre-trained networks.}
 \label{table:cls_conv5_result}
\end{table}

\begin{table}[!b]
\centering
\small
\resizebox{0.75\linewidth}{!}{
 \begin{tabular}{@{}*{3}c@{}}
  \toprule
    \textbf{Feature} &  Top-1 Acc \\
 \midrule
    \texttt{Conv5} & 76.7\\
    \texttt{ConvProj} blob (256,14,14) & 69.7 \\
    \texttt{ConvProj} blob (256,4,4) & 72.4 \\
    \texttt{ConvProj} blob (256,2,2) & 73.3\\
    \texttt{ConvProj} blob (256) & 74.1\\
    \texttt{FC1} (1024) & 71.6\\
    \texttt{FC2} (1024) & 70.0\\
 \bottomrule
 \end{tabular}}
 \caption{Linear classification results on \Cal~\cite{cal256} using different features from  the detection head of \OpenImages~\cite{openimages} pre-trained object detection network.}
 \label{table:cls_inter_result}
\end{table}

\textbf{Intermediate Detection Features} Table \ref{table:cls_inter_result} compares features extracted from different layers in the detection head of the \OpenImages pre-trained object detection network. We present results for classification on the \Cal~\cite{cal256} dataset when a linear classifier is applied to different features, including avg pooled \texttt{Conv5} (2048), \texttt{ConvProj} blob (256,14,14), avg pooled \texttt{ConvProj} blob (256,4,4), avg pooled \texttt{ConvProj} blob (256,2,2), avg pooled \texttt{(ConvProj)} (256), \texttt{FC1} (1024) and \texttt{FC2} (1024) features. We find that \texttt{FC1} is better than \texttt{FC2}. The avg pooled \texttt{(ConvProj)} (256) is better than avg pooled \texttt{ConvProj} blob (256,2,2), which is better than \texttt{ConvProj} blob (256,4,4). Therefore, it is evident that preserving spatial information hurts image classification. Although averaging is an operation which can be learned from a higher dimensional representation (like \texttt{ConvProj} blob (256,14,14)), it is also easily possible to overfit to the training set in a higher-dimensional feature space. We also find that as we approach the \texttt{Output} layer of detection, the performance for image classification deteriorates.

\begin{figure}[t]
  \includegraphics[width=\linewidth]{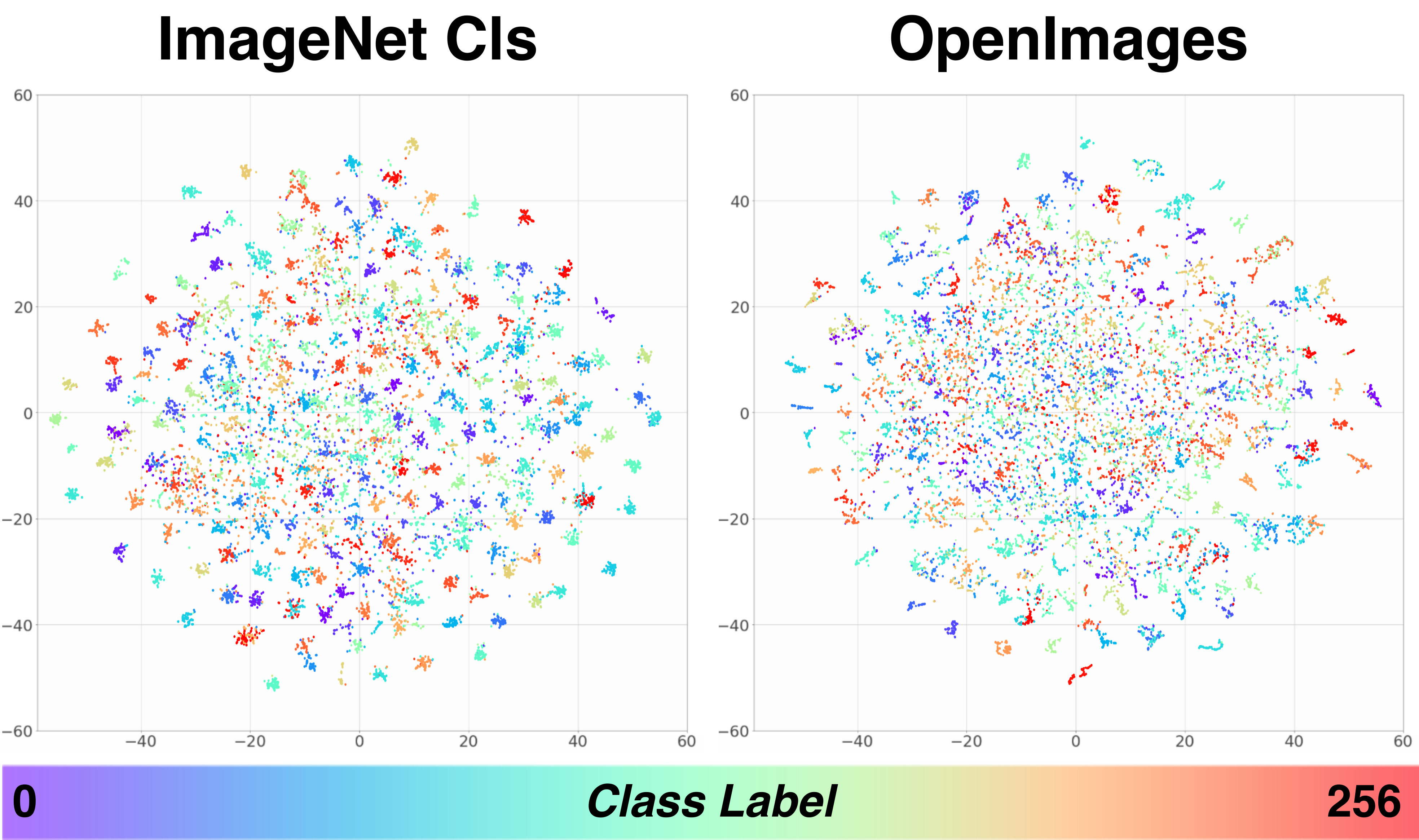}
  \caption{t-SNE~\cite{tsne} visualization of avg pooled \texttt{Conv5} features from \ImageNetCls~\cite{imgnet} and \OpenImages~\cite{openimages} pre-trained networks. The \ImageNetCls features are clustered while \OpenImages features are fragmented.}
  \label{fig:tsne}
\end{figure}

\textbf{Semantic and Feature Analysis}
In Fig \ref{fig:cls_quals} we show the most similar images (in the avg. pooled \texttt{Conv5} feature space, using correlation) for \ImageNetCls and \OpenImages pre-trained networks on \Cal. As can be seen, similar images from \ImageNetCls features can have multiple objects; however for \OpenImages, the most similar image pairs typically match in shape and size. To understand the relationship between \OpenImages and \ImageNetCls features, we perform K-means clustering with different numbers of clusters (from 2 to 1000). Then, given an image pair in the same cluster in an embedding (like \OpenImages), we check if the same image pair belongs to the same cluster in another embedding (like ImageNet) or not. We plot this probability in Fig \ref{fig:feat_subsume}. This plot shows that if features are similar in the \OpenImages space, they are likely to be similar in the \ImageNetCls space; however the converse is not true. Some example images which are close in the \ImageNetCls space but distant in the \OpenImages space are shown in the middle of Fig \ref{fig:cls_quals}. This shows that objects of different scale and similar texture can be close in the \ImageNetCls space but far away in the \OpenImages space. We briefly describe how we define close and distant. An image pair is considered to be close if it is part of the same cluster when the number of clusters is large ($>$ 1000). An image pair is considered to be distant if it not part of the same cluster when the number of clusters is small ($<$ 5). 

We also show the t-SNE~\cite{tsne} visualization of avg pooled \texttt{Conv5} features from \ImageNetCls and \OpenImages pre-trained networks before fine-tuning. We use Barnes-Hut-SNE~\cite{bhsne} and set $perplexity$ and $theta$ to $10.0$ and $0.5$ respectively.  Results in Fig \ref{fig:tsne} show that features from the same class are clustered and close to each other in the \ImageNetCls space; however, \OpenImages features are fragmented.

\subsection{Visualization}

\begin{figure}[t]
  \includegraphics[width=\linewidth]{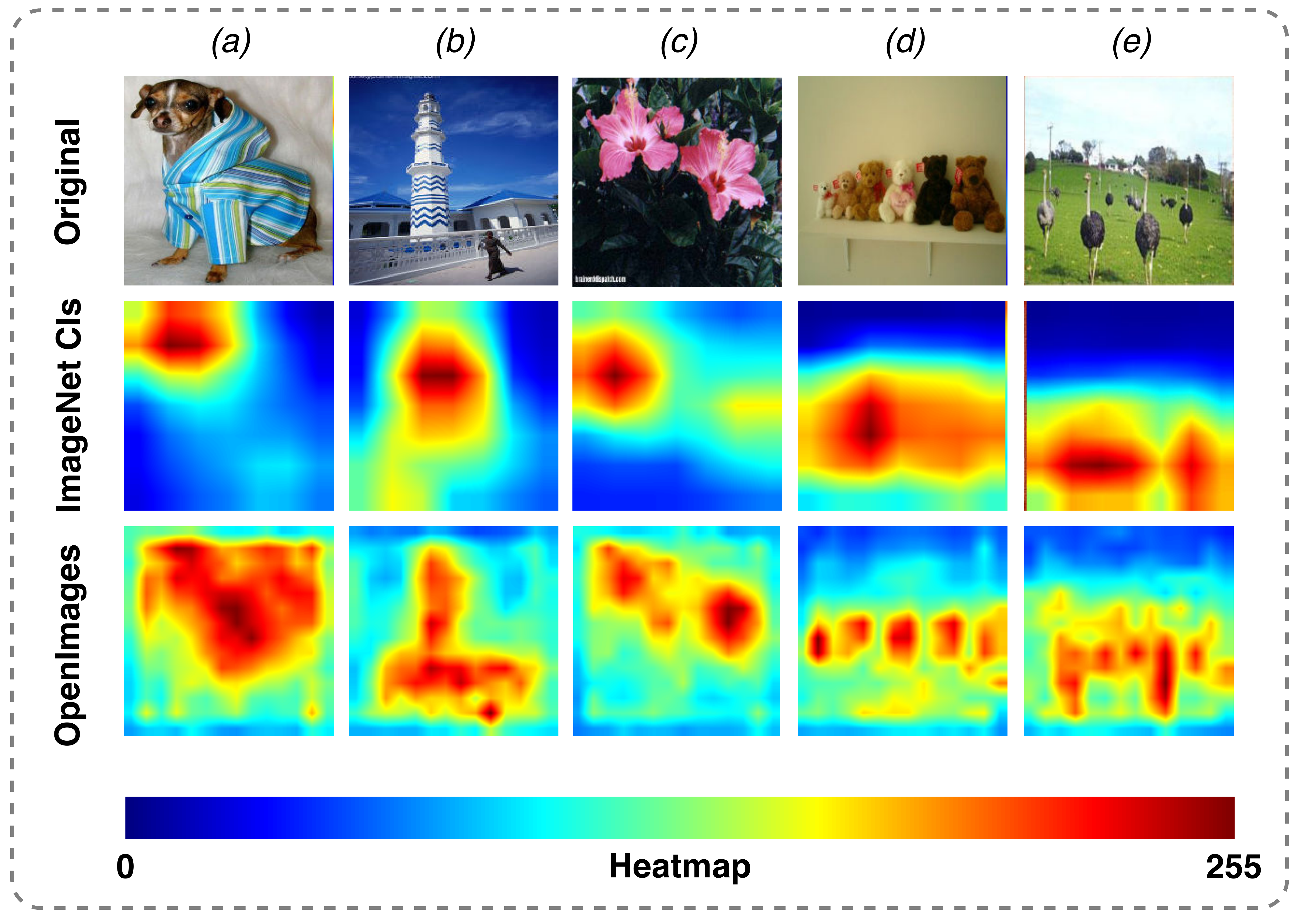}
  \caption{Activation visualization of networks pre-trained on \ImageNetCls~\cite{imgnet} and \OpenImages~\cite{openimages}. Activation maps are averaged across channels, normalized and scaled to range [0, 255], resized to input image size and then colorized. }
  \label{fig:act_vis}
\end{figure}

\textbf{Activation Visualization} The previous quantitative and qualitative analysis suggests considerable differences between networks pre-trained on detection and classification datasets. To illustrate the differences in their internal representations, we first visualize the CNN activations (\texttt{Conv5}) and investigate which part of input images contribute more. As shown in Fig~\ref{fig:act_vis}(a-b), the \ImageNetCls pre-trained activations tend to focus on discriminative parts. On the other hand, \OpenImages pre-trained models emphasize the entire spatial extent of the objects. Moreover, the latter exhibits an instance-level representation, especially when multiple objects are present such as Fig~\ref{fig:act_vis}(c-e).

\begin{figure}[t]
\centering
  \includegraphics[width=0.9\linewidth]{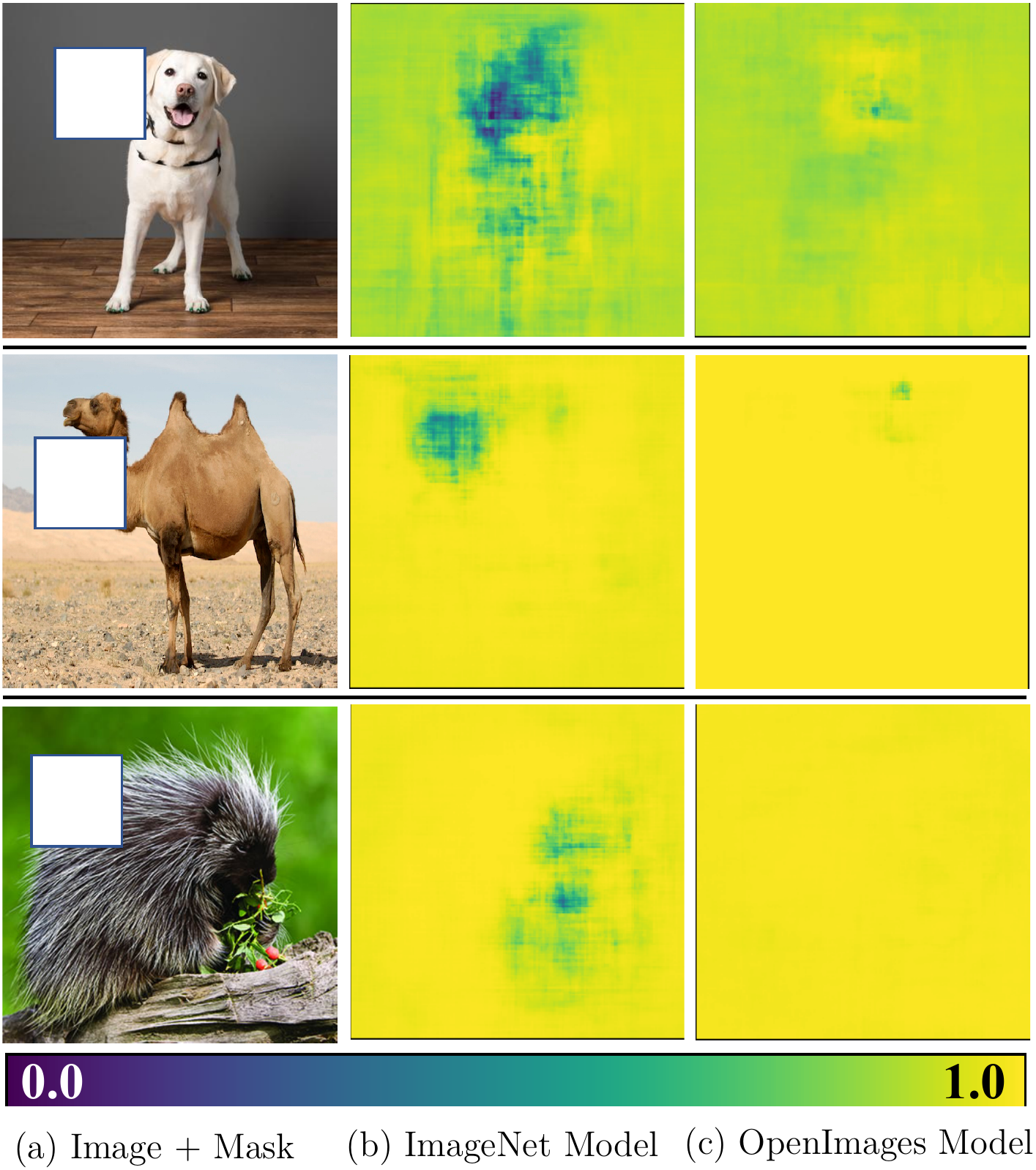}
  \caption{Mask-out visualization. (a) A 60x60 blank mask is shifted over the image. The probability of the correct class at each mask position is shown as a probability map for the \ImageNetCls pre-trained and \OpenImages pre-trained models in (b) and (c) respectively. Unlike the \OpenImages model, the ImageNet-based classifier decides based on specific regions inside the image.}
  \label{fig:maskout_vis}
\end{figure}

\textbf{Mask-out visualization} Besides visualizing activation maps, we further conduct ``Mask-out'' visualization to reveal the relationship between image parts and the final class prediction. Specifically, we shift a blank mask over the input image and measure the output confidence of the correct class. We conduct this experiment on the \Cal dataset. The classification layer for \ImageNetCls and the detection head for \OpenImages is replaced with a linear classification layer. In Fig~\ref{fig:maskout_vis}, we show the classification probability at each pixel assuming that the center of the mask is placed at that location. We can see that for many locations (like the head of the dog or the camel), the classification score of the \ImageNetCls classifier drops to zero, which is not the case for \OpenImages. This is because it relies on the entire spatial extent of an object to make a prediction so the classification score is not sensitive to minor structural changes in the image. However, the \ImageNetCls pre-trained network classifies images based on discriminative parts and when a critical part is masked out, the classification score drops significantly.

\section{Conclusion}

We presented an extensive study on object detection pre-training. When fine-tuning on small detection datasets, we showed that pre-training on large detection datasets is very beneficial when a higher degree of localization is desired. Typically, detection pre-training is beneficial for tasks where spatial information is important such as detection and segmentation, but when spatial invariance is needed, like classification, it can hurt performance. Our feature-level analysis suggests that if detection features of an image are similar, it is likely that their classification features would also be similar while the converse may not hold. Visualization of activations indicates that detection networks focus more on the entire extent of an object while classification networks typically focus on parts. Thus, when minor structural changes are made to an image, detection networks would be robust compared to those trained for classification. 
 \\

\textbf{Acknowledgement} This research was supported in part by the Office of Naval Research (ONR) under Grant N000141612713: Visual Common Sense Reasoning for Multi-agent Activity Prediction and Recognition and the Intelligence Advanced Research Projects Activity (IARPA) via Department of Interior/Interior Business Center (DOI/IBC) contract number D17PC00345. The views and conclusions contained herein are those of the authors and should not be interpreted as necessarily representing the official policies or endorsements, either expressed or implied of ONR, IARPA, DOI/IBC or the U.S. Government.


{\small
\bibliographystyle{ieee}
\bibliography{paper}

\begin{thebibliography}{10}\itemsep=-1pt

\bibitem{tl_factor}
H.~Azizpour, A.~S. Razavian, J.~Sullivan, A.~Maki, and S.~Carlsson.
\newblock Factors of transferability for a generic convnet representation.
\newblock {\em IEEE transactions on pattern analysis and machine intelligence},
  38(9):1790--1802, 2016.

\bibitem{texture_2}
W.~Brendel and M.~Bethge.
\newblock Approximating cnns with bag-of-local-features models works
  surprisingly well on imagenet.
\newblock {\em International Conference on Learning Representations}, 2019.

\bibitem{vocseg3}
L.-C. Chen, G.~Papandreou, I.~Kokkinos, K.~Murphy, and A.~L. Yuille.
\newblock Semantic image segmentation with deep convolutional nets and fully
  connected crfs.
\newblock {\em arXiv preprint arXiv:1412.7062}, 2014.

\bibitem{deeplab}
L.-C. Chen, G.~Papandreou, I.~Kokkinos, K.~Murphy, and A.~L. Yuille.
\newblock Deeplab: Semantic image segmentation with deep convolutional nets,
  atrous convolution, and fully connected crfs.
\newblock {\em IEEE transactions on pattern analysis and machine intelligence},
  40(4):834--848, 2018.

\bibitem{deeplabv3}
L.-C. Chen, Y.~Zhu, G.~Papandreou, F.~Schroff, and H.~Adam.
\newblock Encoder-decoder with atrous separable convolution for semantic image
  segmentation.
\newblock In {\em Proceedings of the European Conference on Computer Vision
  (ECCV)}, pages 801--818, 2018.

\bibitem{tl_cui}
Y.~Cui, Y.~Song, C.~Sun, A.~Howard, and S.~Belongie.
\newblock Large scale fine-grained categorization and domain-specific transfer
  learning.
\newblock In {\em Proceedings of the IEEE conference on computer vision and
  pattern recognition}, pages 4109--4118, 2018.

\bibitem{rfcn}
J.~Dai, Y.~Li, K.~He, and J.~Sun.
\newblock R-fcn: Object detection via region-based fully convolutional
  networks.
\newblock In {\em Advances in neural information processing systems}, pages
  379--387, 2016.

\bibitem{dcnv1}
J.~Dai, H.~Qi, Y.~Xiong, Y.~Li, G.~Zhang, H.~Hu, and Y.~Wei.
\newblock Deformable convolutional networks.
\newblock In {\em Proceedings of the IEEE international conference on computer
  vision}, pages 764--773, 2017.

\bibitem{imgnet}
J.~Deng, W.~Dong, R.~Socher, L.-J. Li, K.~Li, and L.~Fei-Fei.
\newblock Imagenet: A large-scale hierarchical image database.
\newblock In {\em 2009 IEEE conference on computer vision and pattern
  recognition}, pages 248--255. Ieee, 2009.

\bibitem{decaf}
J.~Donahue, Y.~Jia, O.~Vinyals, J.~Hoffman, N.~Zhang, E.~Tzeng, and T.~Darrell.
\newblock Decaf: A deep convolutional activation feature for generic visual
  recognition.
\newblock In {\em International conference on machine learning}, pages
  647--655, 2014.

\bibitem{vis_syn_4}
D.~Erhan, Y.~Bengio, A.~Courville, and P.~Vincent.
\newblock Visualizing higher-layer features of a deep network.
\newblock {\em University of Montreal}, 1341(3):1, 2009.

\bibitem{pascal}
M.~Everingham, L.~Van~Gool, C.~K. Williams, J.~Winn, and A.~Zisserman.
\newblock The pascal visual object classes (voc) challenge.
\newblock {\em International journal of computer vision}, 88(2):303--338, 2010.

\bibitem{texture_1}
R.~Geirhos, P.~Rubisch, C.~Michaelis, M.~Bethge, F.~A. Wichmann, and
  W.~Brendel.
\newblock Imagenet-trained cnns are biased towards texture; increasing shape
  bias improves accuracy and robustness.
\newblock {\em International Conference on Learning Representations}, 2019.

\bibitem{rcnn}
R.~Girshick, J.~Donahue, T.~Darrell, and J.~Malik.
\newblock Rich feature hierarchies for accurate object detection and semantic
  segmentation.
\newblock In {\em Proceedings of the IEEE conference on computer vision and
  pattern recognition}, pages 580--587, 2014.

\bibitem{cal256}
G.~Griffin, A.~Holub, and P.~Perona.
\newblock Caltech-256 object category dataset.
\newblock 2007.

\bibitem{vocseg_data}
B.~Hariharan, P.~Arbel{\'a}ez, L.~Bourdev, S.~Maji, and J.~Malik.
\newblock Semantic contours from inverse detectors.
\newblock In {\em 2011 International Conference on Computer Vision}, pages
  991--998. IEEE, 2011.

\bibitem{vocseg1}
B.~Hariharan, P.~Arbel{\'a}ez, R.~Girshick, and J.~Malik.
\newblock Simultaneous detection and segmentation.
\newblock In {\em European Conference on Computer Vision}, pages 297--312.
  Springer, 2014.

\bibitem{rethinking}
K.~He, R.~Girshick, and P.~Doll{\'a}r.
\newblock Rethinking imagenet pre-training.
\newblock {\em arXiv preprint arXiv:1811.08883}, 2018.

\bibitem{maskrcnn}
K.~He, G.~Gkioxari, P.~Doll{\'a}r, and R.~Girshick.
\newblock Mask r-cnn.
\newblock In {\em Proceedings of the IEEE international conference on computer
  vision}, pages 2961--2969, 2017.

\bibitem{scene_1}
L.~Herranz, S.~Jiang, and X.~Li.
\newblock Scene recognition with cnns: objects, scales and dataset bias.
\newblock In {\em Proceedings of the IEEE Conference on Computer Vision and
  Pattern Recognition}, pages 571--579, 2016.

\bibitem{derek}
D.~Hoiem, Y.~Chodpathumwan, and Q.~Dai.
\newblock Diagnosing error in object detectors.
\newblock In {\em European conference on computer vision}, pages 340--353.
  Springer, 2012.

\bibitem{trans_02}
M.~Huh, P.~Agrawal, and A.~A. Efros.
\newblock What makes imagenet good for transfer learning?
\newblock {\em arXiv preprint arXiv:1608.08614}, 2016.

\bibitem{trimap_1}
P.~Kohli, P.~H. Torr, et~al.
\newblock Robust higher order potentials for enforcing label consistency.
\newblock {\em International Journal of Computer Vision}, 82(3):302--324, 2009.

\bibitem{tl_quoc}
S.~Kornblith, J.~Shlens, and Q.~V. Le.
\newblock Do better imagenet models transfer better?
\newblock {\em arXiv preprint arXiv:1805.08974}, 2018.

\bibitem{trimap_2}
P.~Kr{\"a}henb{\"u}hl and V.~Koltun.
\newblock Efficient inference in fully connected crfs with gaussian edge
  potentials.
\newblock In {\em Advances in neural information processing systems}, pages
  109--117, 2011.

\bibitem{shape_2}
J.~Kubilius, S.~Bracci, and H.~P.~O. de~Beeck.
\newblock Deep neural networks as a computational model for human shape
  sensitivity.
\newblock {\em PLoS computational biology}, 12(4):e1004896, 2016.

\bibitem{openimages}
A.~Kuznetsova, H.~Rom, N.~Alldrin, J.~Uijlings, I.~Krasin, J.~Pont-Tuset,
  S.~Kamali, S.~Popov, M.~Malloci, T.~Duerig, and V.~Ferrari.
\newblock The open images dataset v4: Unified image classification, object
  detection, and visual relationship detection at scale.
\newblock {\em arXiv:1811.00982}, 2018.

\bibitem{cornernet}
H.~Law and J.~Deng.
\newblock Cornernet: Detecting objects as paired keypoints.
\newblock In {\em Proceedings of the European Conference on Computer Vision
  (ECCV)}, pages 734--750, 2018.

\bibitem{fpn}
T.-Y. Lin, P.~Doll{\'a}r, R.~Girshick, K.~He, B.~Hariharan, and S.~Belongie.
\newblock Feature pyramid networks for object detection.
\newblock In {\em Proceedings of the IEEE Conference on Computer Vision and
  Pattern Recognition}, pages 2117--2125, 2017.

\bibitem{coco}
T.-Y. Lin, M.~Maire, S.~Belongie, J.~Hays, P.~Perona, D.~Ramanan,
  P.~Doll{\'a}r, and C.~L. Zitnick.
\newblock Microsoft coco: Common objects in context.
\newblock In {\em European conference on computer vision}, pages 740--755.
  Springer, 2014.

\bibitem{ssd}
W.~Liu, D.~Anguelov, D.~Erhan, C.~Szegedy, S.~Reed, C.-Y. Fu, and A.~C. Berg.
\newblock Ssd: Single shot multibox detector.
\newblock In {\em European conference on computer vision}, pages 21--37.
  Springer, 2016.

\bibitem{fcn}
J.~Long, E.~Shelhamer, and T.~Darrell.
\newblock Fully convolutional networks for semantic segmentation.
\newblock In {\em Proceedings of the IEEE conference on computer vision and
  pattern recognition}, pages 3431--3440, 2015.

\bibitem{vocseg2}
J.~Long, E.~Shelhamer, and T.~Darrell.
\newblock Fully convolutional networks for semantic segmentation.
\newblock In {\em Proceedings of the IEEE conference on computer vision and
  pattern recognition}, pages 3431--3440, 2015.

\bibitem{tsne}
L.~v.~d. Maaten and G.~Hinton.
\newblock Visualizing data using t-sne.
\newblock {\em Journal of machine learning research}, 9(Nov):2579--2605, 2008.

\bibitem{instagram}
D.~Mahajan, R.~Girshick, V.~Ramanathan, K.~He, M.~Paluri, Y.~Li, A.~Bharambe,
  and L.~van~der Maaten.
\newblock Exploring the limits of weakly supervised pretraining.
\newblock In {\em Proceedings of the European Conference on Computer Vision
  (ECCV)}, pages 181--196, 2018.

\bibitem{vis_syn_2}
A.~Mahendran and A.~Vedaldi.
\newblock Visualizing deep convolutional neural networks using natural
  pre-images.
\newblock {\em International Journal of Computer Vision}, 120(3):233--255,
  2016.

\bibitem{vis_syn_nips}
A.~Nguyen, A.~Dosovitskiy, J.~Yosinski, T.~Brox, and J.~Clune.
\newblock Synthesizing the preferred inputs for neurons in neural networks via
  deep generator networks.
\newblock In {\em Advances in Neural Information Processing Systems}, pages
  3387--3395, 2016.

\bibitem{flowers}
M.-E. Nilsback and A.~Zisserman.
\newblock Automated flower classification over a large number of classes.
\newblock In {\em Proceedings of the Indian Conference on Computer Vision,
  Graphics and Image Processing}, Dec 2008.

\bibitem{yolo}
J.~Redmon, S.~Divvala, R.~Girshick, and A.~Farhadi.
\newblock You only look once: Unified, real-time object detection.
\newblock In {\em Proceedings of the IEEE conference on computer vision and
  pattern recognition}, pages 779--788, 2016.

\bibitem{shape_1}
S.~Ritter, D.~G. Barrett, A.~Santoro, and M.~M. Botvinick.
\newblock Cognitive psychology for deep neural networks: A shape bias case
  study.
\newblock In {\em Proceedings of the 34th International Conference on Machine
  Learning-Volume 70}, pages 2940--2949. JMLR. org, 2017.

\bibitem{vis_gradcam}
R.~R. Selvaraju, M.~Cogswell, A.~Das, R.~Vedantam, D.~Parikh, and D.~Batra.
\newblock Grad-cam: Visual explanations from deep networks via gradient-based
  localization.
\newblock In {\em Proceedings of the IEEE International Conference on Computer
  Vision}, pages 618--626, 2017.

\bibitem{sharif2014cnn}
A.~Sharif~Razavian, H.~Azizpour, J.~Sullivan, and S.~Carlsson.
\newblock Cnn features off-the-shelf: an astounding baseline for recognition.
\newblock In {\em Proceedings of the IEEE conference on computer vision and
  pattern recognition workshops}, pages 806--813, 2014.

\bibitem{dsod}
Z.~Shen, Z.~Liu, J.~Li, Y.-G. Jiang, Y.~Chen, and X.~Xue.
\newblock Dsod: Learning deeply supervised object detectors from scratch.
\newblock In {\em Proceedings of the IEEE International Conference on Computer
  Vision}, pages 1919--1927, 2017.

\bibitem{vis_simonyan}
K.~Simonyan, A.~Vedaldi, and A.~Zisserman.
\newblock Deep inside convolutional networks: Visualising image classification
  models and saliency maps.
\newblock {\em arXiv preprint arXiv:1312.6034}, 2013.

\bibitem{snip}
B.~Singh and L.~S. Davis.
\newblock An analysis of scale invariance in object detection snip.
\newblock In {\em Proceedings of the IEEE Conference on Computer Vision and
  Pattern Recognition}, pages 3578--3587, 2018.

\bibitem{rfcn3k}
B.~Singh, H.~Li, A.~Sharma, and L.~S. Davis.
\newblock R-fcn-3000 at 30fps: Decoupling detection and classification.
\newblock In {\em Proceedings of the IEEE Conference on Computer Vision and
  Pattern Recognition}, pages 1081--1090, 2018.

\bibitem{sniper}
B.~Singh, M.~Najibi, and L.~S. Davis.
\newblock Sniper: Efficient multi-scale training.
\newblock In {\em Advances in Neural Information Processing Systems}, pages
  9333--9343, 2018.

\bibitem{vis_deconv}
J.~T. Springenberg, A.~Dosovitskiy, T.~Brox, and M.~Riedmiller.
\newblock Striving for simplicity: The all convolutional net.
\newblock {\em arXiv preprint arXiv:1412.6806}, 2014.

\bibitem{jft}
C.~Sun, A.~Shrivastava, S.~Singh, and A.~Gupta.
\newblock Revisiting unreasonable effectiveness of data in deep learning era.
\newblock In {\em Proceedings of the IEEE international conference on computer
  vision}, pages 843--852, 2017.

\bibitem{bhsne}
L.~Van Der~Maaten.
\newblock Barnes-hut-sne.
\newblock {\em arXiv preprint arXiv:1301.3342}, 2013.

\bibitem{action_0}
X.~Wang, R.~Girshick, A.~Gupta, and K.~He.
\newblock Non-local neural networks.
\newblock In {\em Proceedings of the IEEE Conference on Computer Vision and
  Pattern Recognition}, pages 7794--7803, 2018.

\bibitem{action_3}
X.~Wang and A.~Gupta.
\newblock Videos as space-time region graphs.
\newblock In {\em Proceedings of the European Conference on Computer Vision
  (ECCV)}, pages 399--417, 2018.

\bibitem{sun397}
J.~Xiao, J.~Hays, K.~A. Ehinger, A.~Oliva, and A.~Torralba.
\newblock Sun database: Large-scale scene recognition from abbey to zoo.
\newblock In {\em 2010 IEEE Computer Society Conference on Computer Vision and
  Pattern Recognition}, pages 3485--3492. IEEE, 2010.

\bibitem{action_2}
S.~Xie, C.~Sun, J.~Huang, Z.~Tu, and K.~Murphy.
\newblock Rethinking spatiotemporal feature learning: Speed-accuracy trade-offs
  in video classification.
\newblock In {\em Proceedings of the European Conference on Computer Vision
  (ECCV)}, pages 305--321, 2018.

\bibitem{trans_01}
J.~Yosinski, J.~Clune, Y.~Bengio, and H.~Lipson.
\newblock How transferable are features in deep neural networks?
\newblock In {\em Advances in neural information processing systems}, pages
  3320--3328, 2014.

\bibitem{vis_syn_3}
J.~Yosinski, J.~Clune, A.~Nguyen, T.~Fuchs, and H.~Lipson.
\newblock Understanding neural networks through deep visualization.
\newblock {\em arXiv preprint arXiv:1506.06579}, 2015.

\bibitem{taskonomy}
A.~R. Zamir, A.~Sax, W.~Shen, L.~J. Guibas, J.~Malik, and S.~Savarese.
\newblock Taskonomy: Disentangling task transfer learning.
\newblock In {\em Proceedings of the IEEE Conference on Computer Vision and
  Pattern Recognition}, pages 3712--3722, 2018.

\bibitem{vis_guidedbp}
M.~D. Zeiler and R.~Fergus.
\newblock Visualizing and understanding convolutional networks.
\newblock In {\em European conference on computer vision}, pages 818--833.
  Springer, 2014.

\bibitem{pspnet}
H.~Zhao, J.~Shi, X.~Qi, X.~Wang, and J.~Jia.
\newblock Pyramid scene parsing network.
\newblock In {\em Proceedings of the IEEE conference on computer vision and
  pattern recognition}, pages 2881--2890, 2017.

\bibitem{action_1}
B.~Zhou, A.~Andonian, A.~Oliva, and A.~Torralba.
\newblock Temporal relational reasoning in videos.
\newblock In {\em Proceedings of the European Conference on Computer Vision
  (ECCV)}, pages 803--818, 2018.

\bibitem{vis_cam}
B.~Zhou, A.~Khosla, A.~Lapedriza, A.~Oliva, and A.~Torralba.
\newblock Learning deep features for discriminative localization.
\newblock In {\em Proceedings of the IEEE conference on computer vision and
  pattern recognition}, pages 2921--2929, 2016.

\bibitem{places}
B.~Zhou, A.~Lapedriza, A.~Khosla, A.~Oliva, and A.~Torralba.
\newblock Places: A 10 million image database for scene recognition.
\newblock {\em IEEE transactions on pattern analysis and machine intelligence},
  40(6):1452--1464, 2018.

\bibitem{dcnv2}
X.~Zhu, H.~Hu, S.~Lin, and J.~Dai.
\newblock Deformable convnets v2: More deformable, better results.
\newblock {\em arXiv preprint arXiv:1811.11168}, 2018.

\end{thebibliography}
}

\end{document}